%% file: main.tex
\documentclass[10pt,twocolumn,letterpaper]{article}

\usepackage{cvpr}              


\definecolor{cvprblue}{rgb}{0.21,0.49,0.74}
\usepackage[pagebackref,breaklinks,colorlinks,allcolors=cvprblue]{hyperref}

\def\paperID{10} 
\def\confName{EarthVision}
\def\confYear{2025}

\title{Panopticon: Advancing Any-Sensor Foundation Models for Earth Observation}

\author{
    Leonard Waldmann$^{1,2*}$
    \and
    Ando Shah$^{1*}$
    \and
    Yi Wang$^{2}$
    \and
    Nils Lehmann$^{2,3}$
    \and
    Adam J. Stewart$^{2,3}$
    \and
    Zhitong Xiong$^{2}$
    \and
    Xiao Xiang Zhu$^{2,3}$
    \and
    Stefan Bauer$^{2}$
    \and
    John Chuang$^{1}$
}

\usepackage{makecell}
\usepackage{tabularray}
\usepackage{multirow}
\usepackage{graphicx}
\usepackage{tikz}
\usepackage{enumitem}
\usepackage{caption}
\usepackage[accsupp]{axessibility}  

\newcommand{\cir}[1]{\tikz[baseline]{%
    \node[anchor=base, fill=gray, text=white, draw=darkgray, circle, 
    inner sep=0, minimum width=1em]{#1};}}

\newcommand\blfootnote[1]{%
  \begingroup
  \renewcommand\thefootnote{}\footnote{#1}%
  \addtocounter{footnote}{-1}%
  \endgroup
}

\begin{document}

\twocolumn[
  \begin{@twocolumnfalse}
    \maketitle
    \vspace{-2em} 
    \begin{center}
        $^1$University of California, Berkeley;    
        $^2$Technical University of Munich;
        $^3$Munich Center for Machine Learning

        correspondence to {\tt ando@berkeley.edu}\\
        
        
    \end{center}
  \end{@twocolumnfalse}
  \vspace{1em} 
]

\blfootnote{
    $^*$ Denotes co-first authorship, ordered randomly. Co-first authors will prioritize their names on their resumes/websites\\
    Code: \url{https://github.com/Panopticon-FM/panopticon}
}

\input{0_abstract}    
\input{1_intro}

\input{2_related}

\input{3_methods_v2}
\input{4_Experiments}
\input{6_abl_by_stage}
\input{7_conclusion}
{
    \small
    \bibliographystyle{ieeenat_fullname}
    \bibliography{main}
}

\input{final_sup}

\end{document}

%% file: 0_abstract.tex
\begin{abstract}



Earth observation (EO) data features diverse sensing platforms with varying spectral bands, spatial resolutions, and sensing modalities. While most prior work has constrained inputs to fixed sensors, a new class of any-sensor foundation models able to process arbitrary sensors has recently emerged. Contributing to this line of work, we propose Panopticon, an any-sensor foundation model built on the DINOv2 framework. We extend DINOv2 by (1) treating images of the same geolocation across sensors as natural augmentations, (2) subsampling channels to diversify spectral input, and (3) adding a cross attention over channels as a flexible patch embedding mechanism.
By encoding the wavelength and modes of optical and synthetic aperture radar sensors, respectively, Panopticon can effectively process any combination of arbitrary channels.
In extensive evaluations, we achieve state-of-the-art performance on GEO-Bench, especially on the widely-used Sentinel-1 and Sentinel-2 sensors, while out-competing other any-sensor models, as well as domain adapted fixed-sensor models on unique sensor configurations. Panopticon enables immediate generalization to both existing and future satellite missions, advancing sensor-agnostic EO.

\end{abstract}

%% file: 1_intro.tex
\section{Introduction}
\label{sec:intro}

\begin{figure*}[!t]    
    \centering
    \includegraphics[width=\textwidth]{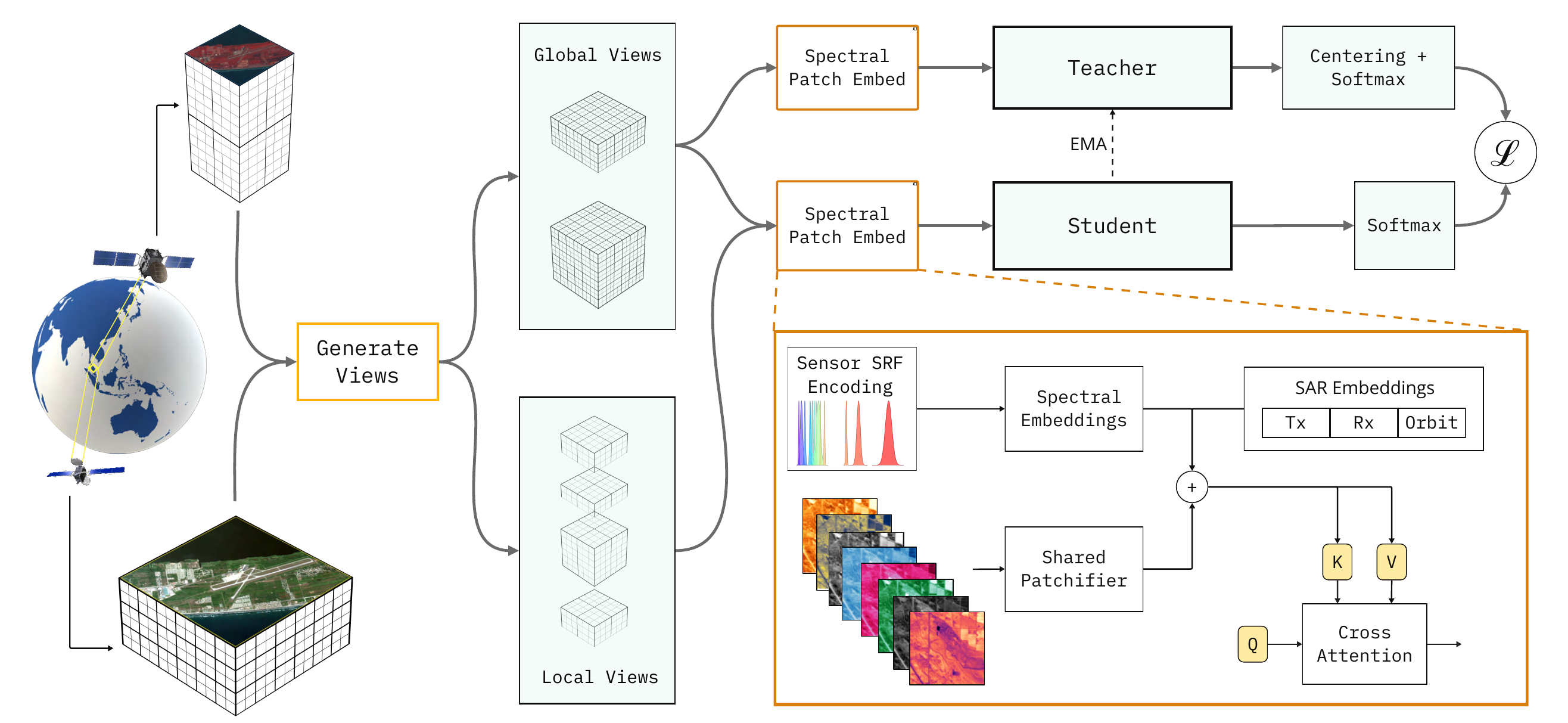}  
    
    \caption{Architecture of Panopticon. The highlighted blocks indicate modifications to the DINOv2 framework. When generating views, images of the same geographic footprint captured by different sensors at different times, scales, etc., are treated as augmentations. The spectral patch embedding block uses a shared patchifier to tokenize incoming views and a cross attention with a single query to fuse tokens within each spatial patch across channels.
    }
    \label{fig:architecture}
\end{figure*}

Earth observation (EO) satellites provide essential data for monitoring our planet, from detecting illegal mining~\cite{suresh2013change} and dark vessels~\cite{paolo2022xview3}, to field delineation~\cite{kerner2024fields} and yield prediction~\cite{rembold2013using}. Recent years have witnessed substantial progress towards remote sensing foundation models (RSFMs) that have adapted computer vision methods designed for RGB imagery to EO data~\cite{reed2023scale, noman2024rethinking, wanyan2023dino, satmae2022}. However, compared to natural images, remote sensing data is remarkably heterogeneous~\cite{ustin2021current, tsokas2022sar, rolf2024mission}. EO platforms generate tremendously varied data across multiple dimensions, including spatial resolution (centimeters to kilometers), spectral characteristics (multispectral, hyperspectral, radar), revisit times (continuous to static), swath widths, preprocessing levels (raw, top of atmosphere, surface reflectance), and sensing mechanisms (active and passive).  While RSFMs began as sensor-specific models~\cite{ma2019deep, satmae2022, jakubik2023foundation, stewart2023ssl4eo, wang2023ssl4eo}, the current trend is towards more flexible models that can deal with a variety of EO data~\cite{xiong2024one, astruc2024anysat, wang2024multi, fuller2023croma, li2024s2mae, tseng2023lightweight}.

Most recently, an emerging class of any-sensor models has begun addressing the fundamental challenge of sensor agnosticism in EO. With only a few examples at the time of writing~\cite{xiong2024neural, prexl2024senpa}, these models process data from arbitrary sensor configurations---including unseen combinations of spectral bands, resolutions, and modalities---without requiring sensor-specific adaptations during inference. They achieve this by decoupling spectral and spatial processing while incorporating knowledge of sensor-specific properties such as spectral response functions (SRFs) and ground sampling distances (GSDs).

In this work, we propose Panopticon, a new any-sensor model (see \Cref{sec:app_why_panopticon} for the etymology of this name). Building upon the DINOv2~\cite{oquab2023dinov2} framework, we make three key modifications for stronger integration of satellite data as a distinct data modality~\cite{rolf2024mission}. 
First, instead of distilling knowledge within a single image, we regard a given geolocation or ``footprint'' as the object of interest and consider snapshots from different sensors of that footprint as augmented views of the same object.
This approach enables us to sample a broad spectrum of natural sensor-to-sensor variation by incorporating images from different channel characteristics, modalities, timestamps, and processing levels. Second, in addition to standard DINOv2 spatial augmentations, we apply a spectral augmentation by subsampling channels of multispectral (MS), hyperspectral (HS), and synthetic aperture radar (SAR) training data. Third, to handle views with varying numbers of channels, we implement a cross-attention over channels with added positional embeddings that encode sensor-specific spectral information into a unified representation.

To evaluate spectral flexibility, we simulate new sensors from HS datasets using spectral convolution~\cite{burggraaff2020biases}, and subject models to reduced spectral and scale information to assess their invariance to these properties, in addition to assessing performance on sensors not encountered during pre-training. We demonstrate that Panopticon also achieves state-of-the-art performance on GEO-Bench~\cite{lacoste2024geo} which comprises of popular sensors such as Sentinel-1 (S1) and Sentinel-2 (S2). To address sensor distribution shifts in fixed-sensor models, we employ patch embedding re-training as parameter-efficient domain adaptation~\cite{hsu2024geospatial, braham2024spectralearth}. A surprising finding is that DINOv2 serves as a remarkably strong baseline in these comparisons, hinting at the nature of future work. Through comprehensive experiments across 23 MS, HS, and SAR datasets, we provide valuable insights into the capabilities and limitations of current approaches to sensor-agnostic learning in EO.

In summary, our contributions are as follows:

\begin{itemize}
    \item We introduce Panopticon, a RSFM capable of building representations for any MS, HS, or SAR sensor, regardless of GSD or spectral characteristics.
    \item While maintaining competitive-to-state-of-the-art performance on established benchmarks, Panopticon outperforms other any-sensor models on more unique sensor configurations that cover important application domains.
    \item We find that spectral progressive pre-training combined with a wider embedding dimension and the use of spectral embeddings in the channel attention module are crucial for optimal performance.
\end{itemize}

\begin{figure*}[th]    
    \centering
    \includegraphics[width=\textwidth]{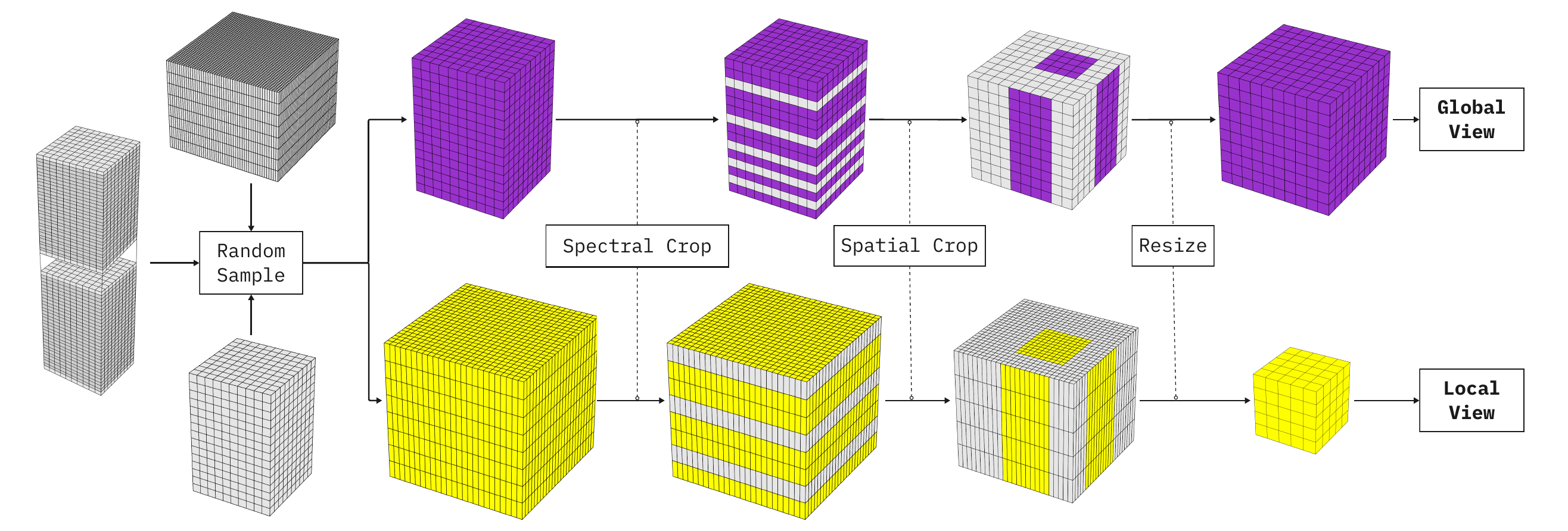}
    \caption{View Generation. Given a footprint, we randomly sample a snapshot performing an implicit spectral, scale, and temporal augmentation. To increase spectral variance, we apply a spectral crop, and following DINOv2, we apply a spatial crop and resize. We omit additional flips and color jittering for simplicity. Local and global views are created analogously with different augmentation parameters.}
    \label{fig:augm}
\end{figure*}

%% file: 2_related.tex
\section{Related work}
\paragraph{Fixed-sensor RSFMs} Approaches vary on how to process different sensors with the same model. USat~\cite{irvin2023usat} and msGFM~\cite{han2024bridging} use sensor-specific patch embeddings feeding into a single encoder, with USat adding positional information to the tokens. SatMAE~\cite{satmae2022}, Galileo~\cite{tseng2025galileo}, and AnySat~\cite{astruc2024anysat} employ more granular approaches with patch embeddings for groups of channels. Video-inspired models like Prithvi~\cite{prithvi} and S2MAE~\cite{li_s2mae_nodate} patchify in spatio-temporal or spatio-spectral dimensions. Another line of work uses separate encoders and aligns the resulting sensor-specific representations through contrastive and reconstruction loss, as in CROMA~\cite{fuller2023croma}, modified contrastive loss, as in IAI-SimCLR~\cite{prexl_multi-modal_2023}, or regularization of covariance matrices, as in DeCUR~\cite{wang_decoupling_2024}.

\paragraph{Any-sensor RSFMs} To the best of our knowledge, two general-purpose any-sensor models have been proposed, both based on the MAE~\cite{he2021_mae} architecture. DOFA~\cite{xiong2024neural} proposes dynamic weight generation with a hypernetwork~\cite{ha2016hypernetworks} based on wavelengths of the input channels for the otherwise fixed-shaped input projection and final layer of the decoder. 
SenPa-MAE~\cite{prexl2024senpa} patchifies each channel separately with the same projection and adds positional, GSD, and spectral embeddings, which are computed from the spectral response function (SRF) of the channel, to the tokens in the encoder and decoder. Concurrent to this work, Copernicus-FM~\cite{wang2025unifiedcopernicusfoundationmodel} extends DOFA with non-spectral modalities for all Copernicus mission sensors.

\paragraph{Sensor views as augmentations} Several RSFMs have utilized the naturally occurring temporal and spectral differences between images of the same footprint as augmentations. In particular, DINO-MC~\cite{wanyan_extending_2024} uses temporal positive pairs, SeCo~\cite{manas_seasonal_2021} incorporates seasonal differences into their branches, and CaCo~\cite{mall_change-aware_2023} defines change-aware positive and negative temporal pairs. On the other hand, DINO-MM~\cite{wang_self-supervised_2022} randomly selects different sensors for each view as a spectral augmentation. Finally, A2MAE~\cite{zhang_2-mae_2024} applies metainfo-aware masking to generate both spectrally and temporally varied tokens.

\paragraph{DINOv2}
DINO~\cite{caron_emerging_2021} matches discrete probability distributions computed by teacher and student networks from differently augmented local and global views of the same image. Student weights are updated with backpropagation while teacher weights are assigned as exponential moving average on the student weights. To prevent collapse, the teacher applies centering and sharpening. 
DINOv2~\cite{oquab2023dinov2} combines DINO with the patch-wise iBOT loss~\cite{zhou_ibot_2022} and scales training. Importantly, a strong DINOv2 ViT-B checkpoint distilled from a ViT-g run is available.

%% file: 3_methods_v2.tex
\begin{table*}[!t]
\centering
\begin{tabular}{lccccccc}
\toprule
 & & & GSD & Footprint & Sensor Modalities & Temporal & Processing \\
Dataset & Locations & Sensors & (m) & (km\textsuperscript{2}) & (\# bands) & Range & Level \\
\midrule
fMoW~\cite{christie2018functional, satmae2022} & 90 K & \makecell{QB, GE, \\ WV, S2} & 0.3--10 & 0.2--25 & \makecell{RGB (3), \\ MSI (4, 8, 12)} & 2002--2020 & L2A \\
MMEarth~\cite{nedungadi2024mmearth} & 1.2 M & S1, S2 & 10 & 1.7 & SAR (8), MSI (12) & 2017--2020 & L1C, L2A \\
SatlasPretrain~\cite{bastani2023satlaspretrain} & 770 K & \makecell{S1, S2,\\ L9} & 10--100 & 25 & \makecell{SAR (2), MSI (9), \\ Thermal (2)} & 2022 & L1C \\
SpectralEarth~\cite{braham2024spectralearth} & 415 K & EnMAP & 30 & 14.7 & HSI (202) & 2022--2024 & L2A \\
\bottomrule
\end{tabular}

\caption{Pretraining Datasets. fMoW refers to the combined fMoW-full and fMoW-Sentinel datasets. Locations refers to the number of unique footprints, where footprint sizes are shown in square kilometers. The parentheses around modalities indicate the number of channels for that modality. MSI = multispectral imager, HSI = hyperspectral imager, QB = QuickBird-1, GE = GeoEye-2, WV = WorldView, S1 = Sentinel-1, S2 = Sentinel-2, and L9 = Landsat~9.}
\label{tab:datasets}
\end{table*}

\section{Method}

We propose Panopticon, which builds on the DINOv2 architecture~\cite{oquab2023dinov2} with changes to view generation and patch embedding, as shown in \Cref{fig:architecture}.

\subsection{Sensor views as augmentation of a footprint}
\label{sec:views}
The DINO framework self-distills knowledge between different augmented views of the same image~\cite{caron_emerging_2021}.
Instead of considering a single image as the object of interest, we consider a geographic footprint as the object of interest and treat images from different sensors as views and therefore augmentations of the same object. Since augmentations define invariances, our model is invariant particularly to spectral configurations but also to GSDs, acquisition times, and processing levels, c.f. \Cref{tab:datasets}.

\subsection{Local and global view generation}
\label{sec:view-gen}
Given a footprint with a set of snapshots $X$ from different sensors of that footprint, let us first generate a local view.
We start by selecting a snapshot $x\in X$ with $C$ channels, therefore implicitly applying a combination of spectral, scale, and temporal augmentations. Since the number of unique sensors in the training dataset is limited in practice, we increase the variance of the spectral information by subsampling channels. In particular, we first sample the number of channels $C'\in\{\min(C,C^l_\text{low}),...,\min(C,C^l_\text{high})\}$ within bounds $C^l_\text{low},C^l_\text{high}$ and then select $C'$ unique channels from $x$. All sampling is done uniformly at random.
As in DINO, we apply a random resize crop resizing to the spatial shape $W^l\times H^l$, c.f. ~\Cref{fig:augm}, as well as additional flips and color jittering.
We repeat this process to generate $n^l$ local views. Analogously, we generate $n^g$ global views with parameters $C^g_\text{low},C^g_\text{high}, W^g, H^g$. 
In correspondence with DINOv2, we set $n^l=4$, $n^g=2$, $H^l=W^l=96$, and $H^g=W^g=224$. To be able to fit a full S2 image into a global crop, we set $C^g_\text{high}=13$ and for maximum flexibility, we set $C^l_\text{low}=1$. Finally, we set $C^l_\text{high}=C^g_\text{low}=4$, c.f. \Cref{sec:add_abl}.

\subsection{Attention over channels as patch embedding}\label{sec:pe}
Let $x\in\mathbb{R}^{C\times H\times W}$ be an input image to the patch embedding (PE) where $C$ is the number of channels, $H$ the height, and $W$ the width. Since $C$ varies across images, we cannot use the standard 2D convolution of ViTs~\cite{dosovitskiy2021imageworth16x16words} as PE. Instead, as in \citet{bao2023channel}, we first patchify and embed each channel with the \emph{same} 2D convolution to $x_p\in\mathbb{R}^{L\times C\times D}$, where $L$ is the number of patches and $D$ the embedding dimension of the backbone. Similarly to \citet{nguyen2023climax}, we then reduce $C$ within each patch by a cross attention with a single learned query $q\in\mathbb{R}^{1\times D}$ and $x_p$ as keys and values to $x_\text{out}\in\mathbb{R}^{L\times D}$. We perform the cross attention on a dimension $D_\text{attn}$, i.e., query, key, and value projections as well as the final projection of the cross attention map to and, respectively, from $D_\text{attn}$. We find that $D_\text{attn}>D$ increases performance, c.f. \Cref{sec:abl_chnattndim}.

To ingest spectral information about the individual channels, we add the following embeddings to $x_p$. 
For optical channels, we choose the standard positional encoding~\cite{vaswani2017attention} at the position equal to the central wavelength $\lambda$ of the channel in nanometers (e.g., 664~nm for red), i.e.
\begin{align}
    \text{PE}_{(\text{pos}, 2i)} &= \sin\left(\omega_i \lambda\right), \\
    \text{PE}_{(\text{pos}, 2i+1)} &= \cos\left(\omega_i \lambda \right),
\end{align}
where $\omega_i = 1/{10000^{2i/D}}$.

We categorize SAR channels into 12 possible categories based on their 3 possible orbit direction (i.e. ascending, descending, unknown/both) and 4 possible transmit and receive polarizations (i.e. VV, VH, HH, HV), providing 12 possible polarization-orbit combinations. We learn embeddings of dimension $D/3$ for each transmit, receiver and orbit variables, and concatenate them to the full embedding vector. This follows known results that SAR data generated across polarizations~\cite{cloude1996review, ulaby1987relating} and orbits~\cite{mahdavi2019effects, ustin2021current} are distinct. Note that we only infuse spectral information but no information on GSD, time, or processing levels.

\subsection{Spectral progressive pre-training}
We found that naively training Panopticon with diverse sensors yields poor performance. Inspired by progressive pre-training \cite{mendieta_towards_2023, xiong2024neural}, we pre-train in two stages with reduced spectral diversity in the first stage, followed by maximizing spectral diversity in the second. In \Cref{sec:abl}, we show that this substantially increases performance.

\subsection{Implementation details}\label{sec:impl_details}

\paragraph{Datasets} We use the datasets presented in \Cref{tab:datasets} with the following rationale:
\begin{itemize}
    \item \textbf{fMoW}, comprising of fMoW~\cite{christie2018functional} and fMoW-Sentinel~\cite{satmae2022}, combines popular optical wavelengths (RGB, S2) and commercial MS sensors - WorldView (WV) 2/3, with high variation in GSD and very high resolution RGB images.
    \item \textbf{SatlasPretrain}~\cite{bastani2023satlaspretrain} pairs S1 SAR, S2 optical, and Landsat~9 optical and thermal sensors across globally diverse footprints.
    \item \textbf{MMEarth}~\cite{nedungadi2024mmearth} provides additional SAR configurations---ascending and descending orbits with VV, VH, HV, and HH polarizations not found in SatlasPretrain. 
    \item \textbf{SpectralEarth}~\cite{braham2024spectralearth} is a hyperspectral dataset with 202 channels, thus provides great spectral diversity over optical wavelengths.
\end{itemize}
Each dataset is standardized with mean and standard deviations computed over its training split. Additional details are available in \Cref{sec:app_dataset}.

\begin{table*}[!t]
\centering
\resizebox{\textwidth}{!}{%
\begin{tabular}{@{}ccccccccc@{}}
\toprule
\multicolumn{1}{l}{} & \multicolumn{2}{c}{\textbf{Commercial Sensors}} & \multicolumn{4}{c}{\textbf{Synthesized Sensors}} & \multicolumn{2}{c}{\textbf{OOD Sensors}} \\
\cmidrule(lr){2-3} \cmidrule(lr){4-7} \cmidrule(l){8-9}
& SpaceNet1 & fMoW-10\% & Corine & Corine & Hyperview & \multicolumn{1}{c}{Hyperview} & TC-10\% & DT-10\% \\
 & WV7 & WV8 & SuperDove & MODIS & SuperDove & \multicolumn{1}{c}{MODIS} & GOES-16 & Himawari \\
& miJacc (\%) $\uparrow$ & Acc (\%) $\uparrow$ & mAP (\%) $\uparrow$ & mAP (\%)$\uparrow$ & MSE $\downarrow$ & \multicolumn{1}{c}{MSE $\downarrow$} & MSE $\downarrow$ & MSE $\downarrow$ \\
\midrule
DINOv2-PE &  90.0 / \textbf{90.6}* & 29.2 / \underline{47.0}* & 66.8 / {80.6}* & 63.6 / \underline{80.7} & 1.043 / 0.340* & 0.639 / \underline{0.335}* & 0.635 & \textbf{0.502} \\
\midrule
Croma-PE & 89.4 & 33.7 & 75.1 & 78.4 & 0.343 & 0.519 & 0.394 & 0.925\\
Softcon-PE & {90.1}  & 32.8 & {77.3} & {80.2} & 0.353 & 0.533 & 0.471 & \underline{0.810}\\
Anysat-PE & 87.7 & 26.2 & 73.4 & 76.6 & \underline{0.326} & 0.519 & 0.449 & 1.406 \\
\midrule
SenPa-MAE & 86.2& 16.9 & 64.2 & 62.7 & 0.366 & 0.355 & - & - \\
DOFA &  89.9 & 42.0 & \underline{81.0} & 80.2 & 0.334 & {0.338} & \underline{0.385} & 1.032 \\
\midrule
Panopticon & \underline{90.3} & \textbf{50.3} & \textbf{85.8} & \textbf{86.2} & \textbf{0.313} & \textbf{0.321} & \textbf{0.315} & 0.963 \\
\bottomrule
\end{tabular}%
}
\caption{Spectral Generalization Performance. Synthesized sensor channels are generated from EnMAP source bands using spectral convolution to indicated target sensors, and tested on the SpectralEarth Corine dataset. Similarly, the Intuition-1 HS sensor is transformed to target bands for the Hyperview dataset. Target bands are generated through spectral convolution for Planet SuperDove's (SD) 8 bands, or MODIS Terra's 16 bands. Out-of-distribution (OOD) sensors refer to weather satellite channels with central wavelengths of 10.4~$\mu$m. `-' indicates the inability of that model to adapt to the dataset. Best results in \textbf{bold}, second-best \underline{underlined}. *Trained re-initialized PE / frozen pre-trained PE using only RGB channels. DT = DigitalTyphoon, TC = TropicalCyclone, miJacc = micro Jaccard index}
\label{tab:spec-gen}
\end{table*}

\begin{table*}[htbp]
\centering
\resizebox{\textwidth}{!}{%
\begin{tabular}{@{}ccccccccccc@{}}
\toprule
         & \multicolumn{6}{c}{\textbf{Classification} (top-1 Accuracy \%) $\uparrow$}             & \multicolumn{4}{c}{\textbf{Segmentation} (miJacc \%) $\uparrow$}               \\ 
\cmidrule(lr){2-7} \cmidrule(l){8-11}
   & \multicolumn{2}{c}{reBEN-10\%} & m-eurosat & m-forestnet & m-so2sat & RESISC45        & m-sacrop         & m-cashew      & m-nzcattle    & m-pv4ger         \\
  & S2          & S1          & S2        & L8          & S2       & RGB           & S2               & S2            & RGB           & RGB              \\ 
\midrule
DINOv2   & 80.1        & -           & \underline{95.5}      & \underline{53.5}        & \underline{60.8}     & \textbf{94.0} & 51.2             & \textbf{65.9} & \underline{92.7}          & \textbf{96.9}    \\ 
\midrule
CROMA    & 79.4        & 70.3        & 91.1      & -           & 53.5     & -             & 48.4             & 44.3          & -             & -                \\
SoftCon  & \textbf{84.3}        & \textbf{80.0}        & 92.2      & -           & 52.1     & -             & \underline{51.3} & 54.5          & -             & -                \\
AnySat   & 76.8           &   64.4         & 87.6      & 50.9        & 42.5     & 65.5          & 39.5             & 38.8          & 92.5          & 92.2             \\
Galileo  & 76.5        & 70.3      & 88.6      & -           & 54.2     & -             & 39.5             & 40.4          & -             & -                \\ 
\midrule
SenPa-MAE & 63.8           & -           & 77.5      & 33.5        & 33.7     & 28.2          & 39.3             & 40.7          & 89.5          & 78.3             \\
DOFA     & 78.8        & 72.0          & 92.9      & 53.2        & 54.2     & \underline{92.0}            & \underline{51.3} & 56.4          & \textbf{92.8} & \underline{96.3} \\ 
\midrule
Panopticon & \underline{83.9} & \underline{78.4} & \textbf{96.4} & \textbf{56.3} & \textbf{61.7} & {90.9} & \textbf{52.6} & \underline{59.3} & {92.6} & 95.2 \\ 
\bottomrule
\end{tabular}%
}
\caption{Classification and segmentation on datasets consisting of well-known sensors from GEO-Bench, reBEN, and RESISC45. `-' indicates the inability of that model to adapt to the dataset. As baseline we report the vanilla pre-trained DINOv2 weights only using RGB channels from the respective dataset, constituting a very strong baseline. Panopticon is in the top 2 models for all but 2 tasks. Best results in \textbf{bold} and second-best \underline{underlined}. L8 = Landsat 8, miJacc = micro Jaccard index}
\label{tab:perf-common}
\end{table*}

\paragraph{Model weights and training} 
We use a ViT-B as backbone with $D=768$ and load the published DINOv2 weights. Since weights for classification and iBOT heads are not provided and training is unstable when randomly initializing them, we pre-train the full DINOv2 weights with the standard patch embedding on fMoW using only the RGB channels and use those head weights as initialization for Panopticon. Finally, we use the attention over channels as PE with $D_\text{attn}=2304$ resulting in 98.1 M parameters for the student, 12.9 M of which belong to the patch embedding.

We train our model in two stages using only fMoW in the first stage and all datasets in the second stage, c.f. \Cref{sec:abl}. Following DINOv2, we artificially define an epoch as 1250 iterations and train for 87.5K iterations or 70 epochs on 16 A100 40GB GPUs with an effective batch size of 1200 for each stage. Further details can be found in the \Cref{sec:app_technical_details}.

%% file: 4_Experiments.tex
\begin{figure*}[t]  
    \centering
    \includegraphics[width=\textwidth]{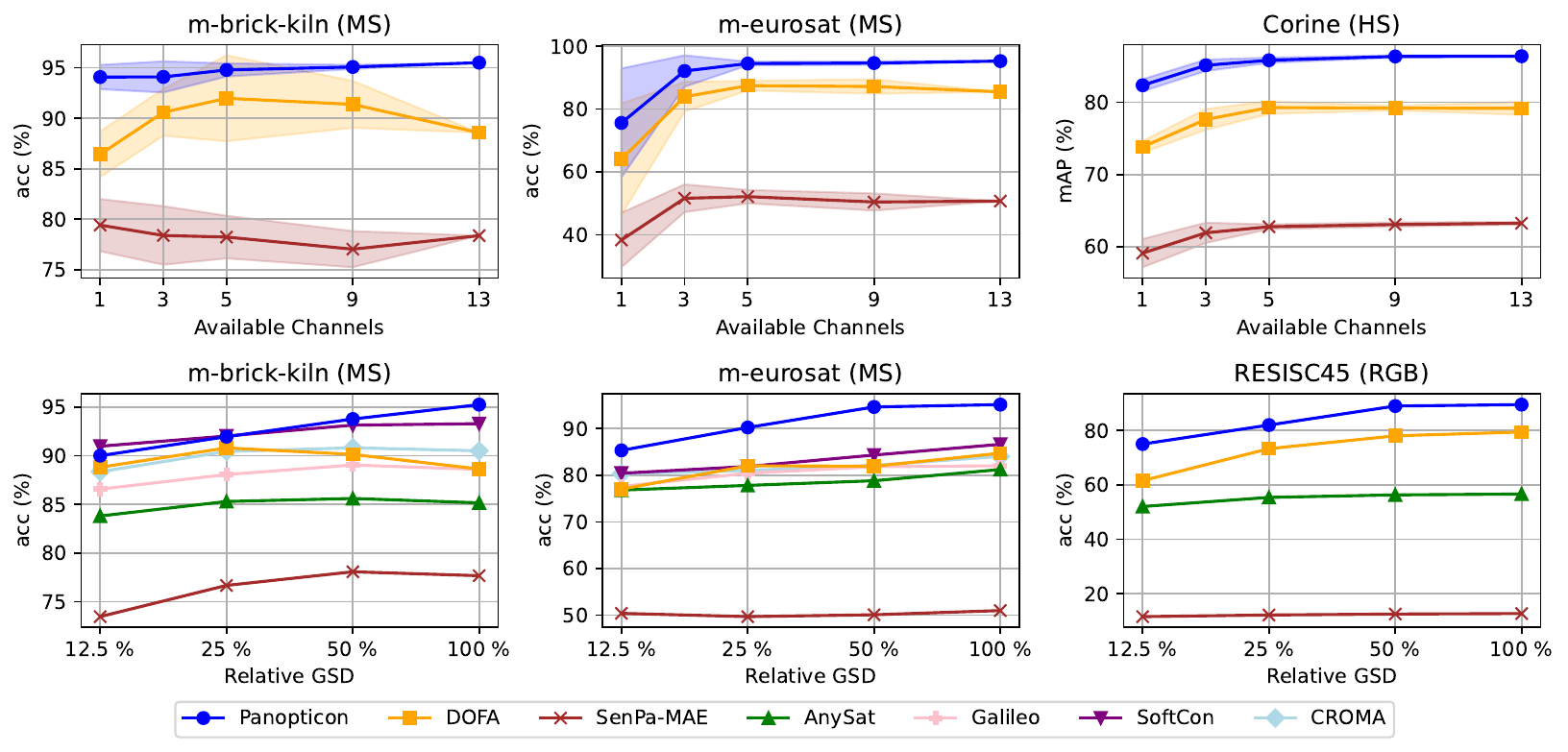}

    \caption{Representation stability under reduction of information. For m-eurosat, m-brick-kiln, and RESISC45, we report top-1 accuracy (acc) from $k$NN classification with $k=20$ and for EnMAP-Corine, we report mean average precision (mAP) from linear probing. Panopticon is shown as a dark blue line in all plots. \textbf{Top:} Learning stable representations under spectral subsampling with 5 different subsets each and corresponding $\pm 1$ standard deviation bands, represented by shaded regions. 
    \textbf{Bottom:} Learning scale-invariant representations by down- and up-sampling both train and test data.
    }
    \label{fig:representation-stability}
\end{figure*}

\section{Evaluation}\label{sec:eval}
We evaluate our model against DINOv2, fixed-sensor models, and any-sensor models in linear probing and $k$NN settings for 11 classification, 7 segmentation, and 4 regression downstream tasks. We evaluate sensor generalization in two specific ways: a) the ability to generalize to unseen sensor configurations, and b) the ability to generate representations that are scale and spectrally invariant. Moreover, we test performance on commonly utilized EO benchmarks such as GEO-Bench~\cite{lacoste2024geo}. All evaluation is conducted on {ViT-Base} backbones.
Additional details on dataset processing and evaluation protocols used throughout this section can be found in \Cref{tab:datasets} and \Cref{sec:app_technical_details}.

\subsection{Generalization to unique sensors}

We evaluate Panopticon against state-of-the-art (SOTA) fixed-sensor models on datasets with sensor configurations that were not explicitly encountered during pre-training. Given the limited availability of open-access data and benchmarks from commercial sensors~\cite{rufin2025enhance}, we employ spectral convolution over HS bands to simulate existing sensors with different spectral characteristics; additional details in the \Cref{sec:std}. Using the SpectralEarth-Corine dataset~\cite{braham2024spectralearth}, we transform the HS bands to correspond to Planet's SuperDove sensor (8 bands)~\cite{planetSuperDove2025, tu2022radiometric} and MODIS Terra sensor (16 bands)~\cite{savtchenko2004terra}, creating the \cir{1}~Corine-SD and \cir{2}~Corine-MODIS datasets, respectively. We apply the same transformation to the Hyperview~\cite{9897254} HS dataset, mapping it to  \cir{3}~SuperDove and  \cir{4}~MODIS bands. Finally, we employ the wind speed estimation tasks \cir{5}~TropicalCylone~\cite{maskey2022tropical} and \cir{6}~DigitalTyphoon~\cite{kitamoto2023digital}, useful for their extreme spectral distribution shifts with wavelengths of 10.4~$\mu$m. 
To adapt fixed-sensor models to new sensors, we re-train a 2D convolution as patch embedding while keeping the backbone frozen, denoted by the suffix ``-PE''. This approach has shown to be a parameter-efficient domain adaptation strategy for EO foundation models~\cite{hsu2024geospatial}. This allows an otherwise fixed-sensor model to adapt to datasets originating from other sensors.

Additionally, we include benchmark datasets that contain two commercial sensors: \cir{7} SpaceNet~1~\cite{etten2018spacenet} which uses the WV2 sensor, and the \cir{23}~fMoW dataset that includes only the WV2+WV3 sensors (10\%).

\Cref{tab:spec-gen} summarizes the results. While the domain-adapted fixed-sensor models work well and often outperform previous any-sensor models, Panopticon almost uniformly excels at these tasks, in most cases by a substantial margin. 

\subsection{Spectral invariance}
Inspired by \citet{reed2023scale}, we evaluate spectral invariance by progressively reducing the number of available channels as illustrated in \Cref{fig:representation-stability} (top). 
We employ $k$-nearest neighbors ($k$NN) classification on the datasets \cir{8}~EuroSAT~\cite{helber2019eurosat} and \cir{9}~Brick Kiln~\cite{lee2021scalable}, both with their GEO-Bench modifications~\cite{lacoste2024geo} and use uniformly subsampled channels. For the HS \cir{10}~EnMAP-Corine dataset~\cite{braham2024spectralearth}, we use linear probing (LP) and perform binned channel subsampling.
For each setting, we sample 5 subsets and report mean and standard deviation of the task metric. Note that sampled channels are identical for all models. 
We only evaluate on any-sensor models since we are interested in representation quality without adjusting the backbone.

Panopticon retains the ability to generate high quality representations across almost the entire range of spectral subsampling, exhibiting high spectral invariance within these tasks. Of particular note is Panopticon's reduced variance across subsampled channels indicating a higher level of channel, and therefore spectral invariance. Both DOFA and SenPa-MAE tend to retain stable performance across the range (albeit with higher variance), suggesting that the pre-training paradigms of any-sensor models likely contribute to this behavior. Additionally, we evaluate Panopticon for its cross-sensor generalization ability and the results can be found in \Cref{sec:cross-sens}.

\subsection{Scale invariance}

We evaluate scale invariance through $k$NN classification across two multispectral and one RGB dataset, as presented in \Cref{fig:representation-stability} (bottom). The MS datasets include \cir{8}~EuroSAT~\cite{hsu2024geospatial} and \cir{9}~Brick Kiln~\cite{lee2021scalable}, both modified according to GEO-Bench specifications~\cite{lacoste2024geo}. For RGB assessment, we utilize the \cir{11}~RESISC45~\cite{cheng2017remote} dataset since it exhibits a large variance in GSDs natively, which we further augment. The MS datasets feature fixed GSDs of 10~m. Similarly to the protocol established by \citet{reed2023scale}, we downsample both train and test sets
to progressively coarser GSDs to assess model robustness to spatial resolution degradation. 

For the MS datasets, Panopticon's representations remain very stable across the GSD range and perform better than the competition. 

\subsection{Performance on popular sensors}

We compare existing fixed- and any-sensor models on the 6 classification and 6 segmentation datasets from GEO-Bench. We substitute \cir{12}~BigEarthNet-S2 and \cir{13}~BigEarthNet-S1 for reBEN, as it is upgraded with more robust splits and lower label noise~\cite{clasen2024reben}. These benchmark tasks represent a wide range of real-world downstream tasks on S1, S2, Landsat~8, RGB, and RGBN sensors. We include \cir{11}~RESISC45~\cite{cheng2017remote} as an additional RGB land cover classification task, as well as an additional 9 tasks from \cir{14}--\cir{22}~GEO-Bench~\cite{lacoste2024geo}. An abridged summary of results is shown in \Cref{tab:perf-common}. The complete GEO-Bench tables, along with the descriptions of the tasks, can be found in \Cref{sec:app_additional_results}.

Panopticon exhibits SOTA results on most tasks, and is competitive on the rest, showing that the model can also perform well on popular open access sensors.

%% file: 6_abl_by_stage.tex
\section{Ablation studies}\label{sec:abl}

\begin{table*}[htbp]
    \centering
    \begin{tabular}{lllrrrr}
        \toprule
         &  &  & $\text{Avg} \uparrow $ & $\text{MS}_\text{acc}\uparrow$ & $\text{SAR}_\text{acc}\uparrow$ & $\text{Sim}_\text{mAP}\uparrow$ \\
         \midrule
        Stage 1 & Default & fMoW & \textbf{83.5} & \textbf{92.2} & 75.8 & 79.6 \\
         \cmidrule(lr){2-7} 
         & Datasets   & MME &  81.7 & 86.4 & \textbf{81.7} & 80.8 \\
         &       & SP &  81.2 & 84.4 & 77.5 & \textbf{82.1} \\
         &      & fMoW, MME & 78.3 & 82.0 & 80.1 & 79.2 \\
         &       & fMoW, SP, MME, SE &  73.9 & 79.1 & 74.4 & 76.8 \\
        \midrule
        Stage 2 & Default & Fine PE, multi-view & \textbf{86.8} & \textbf{93.1} & {82.2} & \textbf{83.4} \\
         \cmidrule(lr){2-7} 
         & PE & None & 84.8 & 91.5 & 81.2 & 80.7 \\
         &  & Coarse & 85.1 & 91.0 & 82.1 & 80.8 \\
         &  & Fine-std & 85.9 & 91.7 & \textbf{82.9} & 80.4 \\
         \cmidrule(lr){2-7} 
         & Views & Single & 82.2 & 84.2 & 78.8 & 83.0 \\
        \bottomrule
    \end{tabular}
    \caption{Ablations of major design decisions of Panopticon by pretraining stages. The metrics are averages in \% over several tasks indicating performance on MS, SAR, and non-standard sensors simulated by channel subsampling. Best results within each stage in \textbf{bold}. SP = SatlasPretrain, SE = SpectralEarth, and MME = MMEarth.}
    \label{tab:abl1}
\end{table*}

We perform ablation studies on both pre-training stages following the configuration of \Cref{sec:impl_details} but reducing the epochs to 30 with an effective batch size of 300. To evaluate performance over MS, SAR, and non-standard sensors with a smaller compute budget than in \Cref{sec:eval}, we define the following metrics: i)~{$\text{MS}_\text{acc}$}: average $k$NN accuracy on m-eurosat with and without RGB channels, ii) {$\text{SAR}_\text{acc}$}: average accuracy of $k$NN and linear probing on Eurosat-sar~\cite{wang2023feature}, iii) {$\text{Sim}_\text{mAP}$}: average mAP across two sensors simulated from HS Corine data by channel subsampling,
and iv) {$\text{Avg}$}: average across all individual tasks. Details on metrics can be found in \Cref{sec:app_technical_details} and additional ablations studies in \Cref{sec:add_abl}. Our key findings are that spectral progressive pretraining, a larger dimensions in channel attention relative to the backbone, and spectral positional embeddings are crucial for good performance.

\paragraph{Progressive pre-training}
We ablate the dataset composition of our pre-training in \Cref{tab:abl1}. We can see that training with multiple datasets and, thus, more spectrally diverse inputs leads to suboptimal results. 
Note that, e.g., training with fMoW and MMEarth performs worse on all metrics than training on only  MMEarth.
This motivates our two-stage pre-training approach. Following \Cref{tab:abl1}, we only train on fMoW in the first stage and increase spectral diversity by training on all datasets in the second stage. While possible, we found no benefits of using more than these two stages.

\paragraph{Channel attention architecture}\label{sec:abl_chnattndim}
We sweep different number of heads $n_h$ and embedding dimensions $D_\text{attn}$ in the channel attention of our PE on stage 1 using two learning rates with 15 epochs each. Results can be seen in \Cref{tab:abl2}. Most notably, using $D_\text{attn}\geq1536$ improves performance significantly compared to $D_\text{attn}=768$. The influence of $n_h$ is less profound. Note that increasing $D_\text{attn}$ comes with substantially increasing the number of parameters as can be seen in the second row of \Cref{tab:abl2}.

\begin{table}[htbp]
    \centering
    \begin{tabular}{lrrrrr}
        \toprule
        $D_\text{attn}$ &  &  768  &  1536 &  2304 &  3072 \\
        \# params (in M) & & 1.9  & 6.2  & 12.9  & 21.9  \\
        \midrule
        $n_h$ & 12     &  77.7 &  82.4  &  83.9 &   - \\
        & 16     &  77.8 &  84.9  &  \textbf{85.8} &  81.5 \\
        & 24     &   -   &  83.8  &  \underline{85.5} &  84.4 \\
        & 32     &   -   &  -     &  - &  82.1 \\
        \bottomrule
    \end{tabular}
    \caption{Ablation of the channel attention architecture. $\text{MS}_\text{acc}$ metric reported when changing the number of heads $n_{h}$ in the rows and embedding dimension $D_\text{attn}$ in the columns. The number of parameters of the channel attention module is shown in millions in the second row. Best results in \textbf{bold}, second-best \underline{underlined}.}
    \label{tab:abl2}
\end{table}

\paragraph{Sensor views as augmentations}
When generating a view of a given footprint, Panopticon samples a sensor image from all available images of that footprint as augmentation. In \Cref{tab:abl1}, we present an ablation (`Single') were we generate all views of a given footprint from a single image. This substantially reduces performance.

\paragraph{Spectral embedding}
In \Cref{tab:abl1}, we compare our proposed embedding of encoding individual wavelengths and SAR configurations (`Fine-PE') against a coarse embedding that only differentiates between SAR and optical channels (`Coarse'), no embedding at all (`None'), and an embedding that additionally encodes the standard deviation of the SRF of optical sensors (`Fine-std'), see \Cref{sec:std} for details.
For optical sensors, having fine-grained spectral information substantially improves performance while adding information about the standard deviation does not seem to provide a benefit. For SAR, the coarse embedding may be sufficient. However, note that none of the downstream datasets include SAR data other than a single transmit and receiver polarization (VV+VH), limiting our ability to validate this outcome. For additional insights into design decisions, please refer to \Cref{tab:abl_cumulative} and \Cref{sec:design-dec}.

%% file: 7_conclusion.tex
\section{Discussion and conclusion}

In this work, we introduced Panopticon, a flexible any-sensor foundation model that processes data from arbitrary sensor configurations without sensor-specific adaptations. By extending DINOv2 with multi-sensor view generation, spectral subsampling, and cross-attention over channels, our approach achieves state-of-the-art performance on standard benchmarks while demonstrating superior generalization across sensor configurations. Our rigorous evaluation tested the model's sensor-agnostic capabilities through reduced spatial and spectral information tests and cross-sensor generalization. Our findings show that Panopticon consistently outperforms existing fixed-sensor and any-sensor models in developing truly sensor-agnostic representations. Tests on datasets outside typical training distributions revealed that most models can adapt to novel sensor modalities, though with varying success rates. Notably, DINOv2 with appropriate domain adaptation techniques proved a surprisingly strong baseline for RGB applications and certain MS use-cases.

Our evaluation across 23 diverse tasks represents significant progress in validating any-sensor capabilities, though limitations remain. We did not explore temporal invariance or equivariance—a critical direction for future work—nor fully investigate incorporating channel bandwidth information and using spectral convolution as augmentation strategies. Additionally, the absence of evaluation tasks utilizing diverse SAR channel combinations limited our ability to comprehensively test the model's SAR generalization capabilities.

While Panopticon explicitly models channel-level sensor characteristics, it relies on view augmentations to learn other important EO invariances related to capture time, processing levels, and ground sampling distance. Though we specifically tested and validated the latter, the other properties proved challenging to isolate and evaluate. Our results suggest that combining explicit spectral modeling with diverse augmentation strategies offers a promising path toward truly sensor-agnostic Earth observation.

The development of robust any-sensor models is constrained by the scarcity of machine learning-ready data from diverse satellite platforms, impeding equitable planetary monitoring \cite{rufin2025enhance}. As remote sensing platforms proliferate, frameworks like Panopticon that generalize across sensor configurations will become increasingly valuable for maximizing scientific and societal benefits. Future work should incorporate additional sensor-specific inductive biases, expand temporal modeling capabilities, and develop comprehensive cross-sensor evaluation benchmarks.

\newpage
\section*{Acknowledgments}

This material is based upon work supported by the National Science Foundation under Grant No.\ DGE- 2125913. The authors gratefully acknowledge the computing time provided on the high-performance computer HoreKa by the National High-Performance Computing Center at KIT (NHR@KIT). This center is jointly supported by the Federal Ministry of Education and Research and the Ministry of Science, Research and the Arts of Baden-Württemberg, as part of the National High-Performance Computing (NHR) joint funding program\footnote{\url{https://www.nhr-verein.de/en/our-partners}}. HoreKa is partly funded by the German Research Foundation (DFG). The authors are also grateful to Prof.\ Joshua Blumenstock, and the Berkeley High Performance Computing (Savio) department for the provisioning of computing infrastructure. 

%% file: final_sup.tex
\setcounter{section}{0} 
\renewcommand{\thesection}{\Alph{section}} 
\setcounter{page}{1}
\renewcommand{\figurename}{Supp. Fig.}
\setcounter{figure}{0} 


\maketitlesupplementary
\label{sec:supp}

\section{Why ``Panopticon"?}\label{sec:app_why_panopticon}

The idea of the \textit{panopticon} was first suggested by the philosopher Jeremy Bentham~\cite{bentham1791panopticon} as an ideal model for efficient prison design, where a single person could watch over an entire prison. Michel Foucault later reinterpreted it as a powerful metaphor for repressive systems of power, control and surveillance in modern societies~\cite{foucault2020panopticism}, something that is perhaps even more relevant today with the proliferation of digital surveillance technologies.

We are well aware of the term's loaded history and controversial connotations; so why choose it? We want to co-opt this term and flip its meaning, using it as a metaphor for systems that can keep a watchful and benevolent eye over our planet. Instead of surveilling people, Panopticon(s) can observe Earth itself---its changing landscapes, ecosystems, and climate patterns.

The beauty of our model is that, like the original panopticon concept, it provides comprehensive visibility from a single vantage point. But unlike Bentham's prison design, our goal is not control and fear, but rather understanding and monitoring. There is also a technical parallel that we find fitting: the original panopticon was designed to see everything from a central position, regardless of where attention was directed. Similarly, our model can ``look'' through any sensor configuration without needing specific adaptations---a sensor-agnostic vision that mirrors the all-seeing nature of the conceptual panopticon, but repurposed for planetary good.

\begin{figure}[ht]    
    \centering
    \includegraphics[width=\columnwidth]{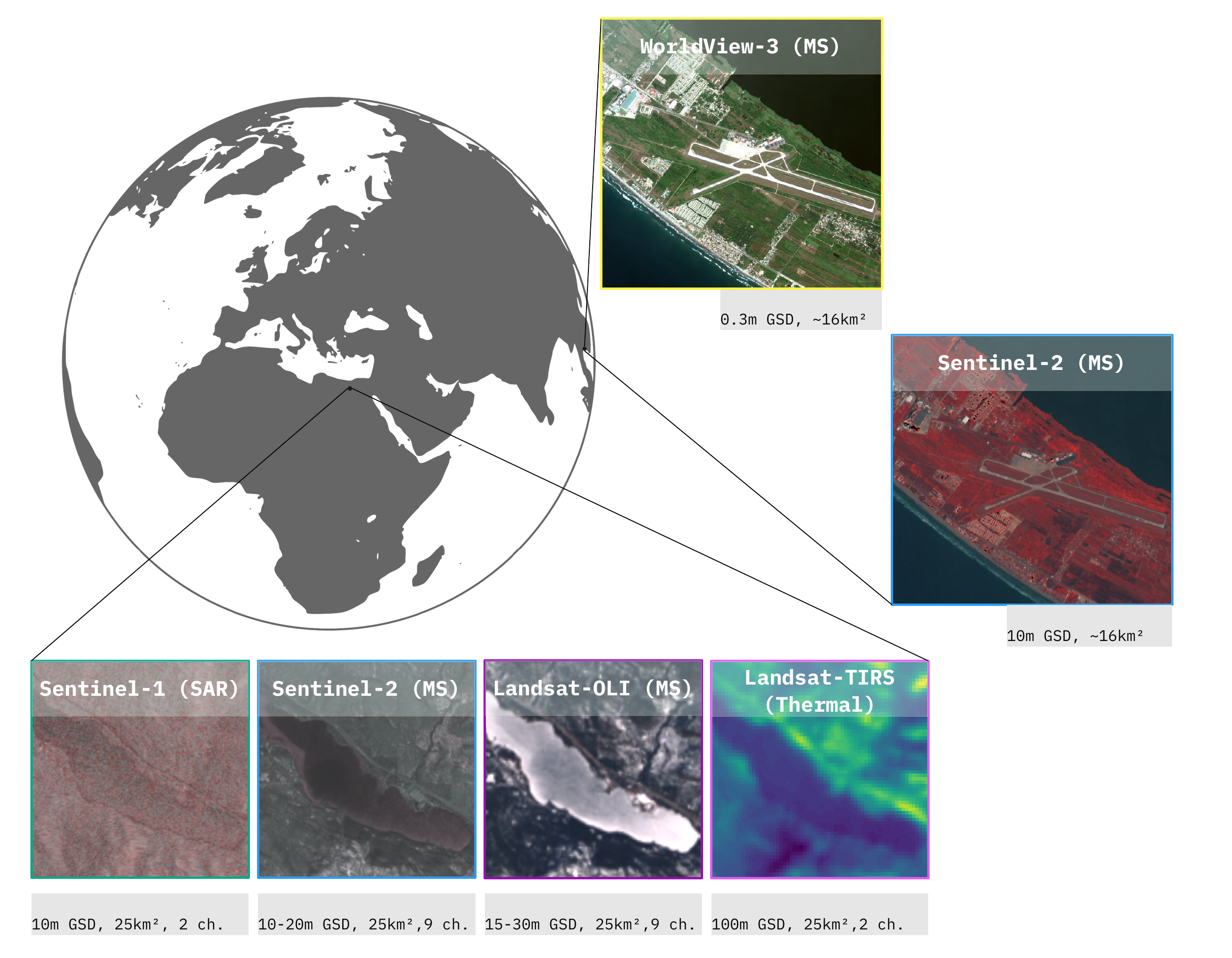}
    \caption{Snapshots: examples of snapshots taken from distinct footprints. Our pretraining dataset consists of various sensor modalities, channels, GSDs, scales, and timesteps acquired from across the Earth's surface. Different channels, GSDs, and footprint sizes provide information about different attributes of the geospatial objects. Note: some images are shown in false colors to enable mapping from non-visible spectra.}
    \label{fig:snapshots}
\end{figure}

\section{Code and data availability}\label{sec:app_code_and_avail}

All data loaders and model code will be contributed to the TorchGeo library~\cite{stewart2022torchgeo} for reproducibility and ease of future experimentation. 

\begin{figure*}[t]  
    \centering
    \includegraphics[width=\textwidth]{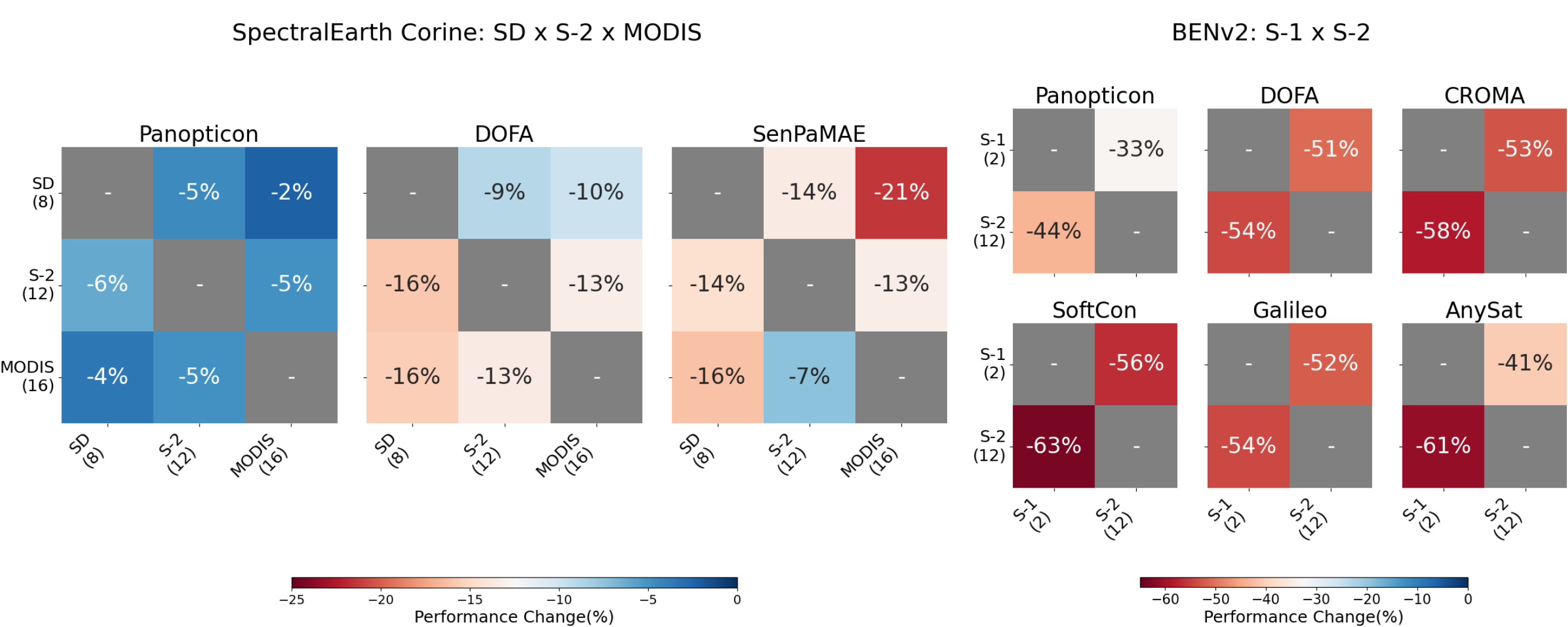}
    \caption{Cross-sensor invariance. In addition to train/val/test splits, we also split datasets across sensors to explicitly test sensor invariance. $y$-axes on the heatmap represent training sensors and $x$-axes, test sensors. The diagonals represent same-sensor for train and test, and are grayed-out, while off-diagonal elements represent cross-sensor results with values expressed as percentage differences from the diagonals. This enables visualization of the change in cross-sensor performance relative to same sensor, expressed in (negative) percentages. 
    \textbf{Left}: splits are across synthesized sensors from the HS EnMAP sensor using spectral convolution - MODIS, Sentinel-2 and Planet SuperDove, for any-sensor models. \textbf{Right}: reBEN; splits are implemented across Sentinel-1 and Sentinel-2 sensors, for all any-sensor and many-sensor models, except SenPaMAE which cannot process SAR imagery. Panopticon consistently outperforms other any-sensor models. Note that the value ranges differ for the two sub-figures.}
    \label{fig:cross-sens}
\end{figure*}

\section{Datasets}\label{sec:app_dataset}

\subsection{Pre-training datasets}

We organize the four pretraining datasets as shown in Table 1 of the main text by geographical footprint, where a footprint is defined by images having an exact geo-reference match. During pretraining, from each footprint, we randomly sample a number of ``snapshots'' as shown in Fig.~\ref{fig:snapshots}.

\paragraph{fMoW}

fMoW~\cite{christie2018functional} consists of data from 4 MS satellite sensors, QuickBird-1 and GeoEye-2 having 4 channels (RGB, NIR), while WorldView-2 and -3 have 8 channels, with GSDs typically ranging from 1--2~m. Additionally, the dataset also includes pansharpened versions of these same images in RGB which have a GSD \textless\,1~m. This dataset was chosen for its global spatial coverage, wide spectral coverage\footnote{including ``non-standard'' bands, i.e. those with spectral coverage outside those of the popular open-data sources such as Landsat and Sentinel series.} and very low GSD values, along with extensive functional coverage of human modified land cover types. Moreover, this dataset also represents a large variance in time of capture, off-nadir angles, both of which affects illumination of targets.
Since spatial footprints were available for all images, we generate an geographically indexed version of this dataset, which will be released with the rest of the code.
Additionally, we remove images greater than 1024~px, which are typically the pansharpened RGB images, to reduce memory overheads.

\paragraph{fMoW-Sentinel}

fMoW-Sentinel~\cite{satmae2022} was created to be an exact copy of the locations captured by fMoW, but with Sentinel-2 imagery. We created a combined dataset from fMoW and fMoW-Sentinel by indexing by footprint and sensor type. Together, these two datasets capture surface properties from five separate sensor platforms between 2002 and 2022, providing a lot of natural variation for a given footprint. Finally, the footprints of each image vary tremendously, from 0.2 to 25~km\textsuperscript{2}, providing a large range of features at different scales. This combined dataset consists of 89,666 unique footprints, where each footprint can have dozens of images across these sensors.

\paragraph{MMEarth}

Multi-modal Earth (MMEarth)~\cite{nedungadi2024mmearth} was released as a paired dataset of multiple modalities that include Sentinel-1 SAR, Sentinel-2 MS, elevation, and other paired modalities. Most importantly, the Sentinel-2 data included multiple processing levels (L1C and L2A), and Sentinel-1 data was captured in all 4 polarization combinations (VV, VH, HH, HV) and in both orbits (ascending, descending). This data was primarily included to model the effects of polarization and orbit for SAR data and processing levels for optical. Moreover, this extensive dataset comprises of 1,239,937 unique footprints equitably distributed according to land cover types, each of which providing a pair of S1 and S2 images.

\paragraph{SpectralEarth}

SpectralEarth~\cite{braham2024spectralearth} is the largest open source hyperspectral dataset available at the time of writing comprising of 450 K patches sampled globally by the EnMAP satellite~\cite{guanter2015enmap}, made available by the German Aerospace Center (DLR). This dataset additionally provides four downstream benchmark tasks using data from the same sensor, but utilizing non-overlapping patches, separate from pretraining. This dataset was included for its rich spectral diversity, enabling the model to learn HS characteristics and simulate any arbitrary multispectral band. SpectralEarth provides 415 K unique footprints, each of which provides a single HS image of 202 bands.

\paragraph{SatlasPretrain}
SatlasPretrain consists of 30~TB of imagery across Sentinel-1, Sentinel-2, Landsat-9, and NAIP sensors. We utilize only the first three, as NAIP geographic coverage is limited to the United States and spectrally consists of only RGB and NIR bands. We created a unified indexed dataset comprising of 768,800 unique footprints, where each footprint can contain up to 3 sensor images taken across 2022. It is also the only large pretraining dataset to contain thermal images from the Landsat 9-TIRS sensor.

\subsection{Evaluation datasets}

The utilize the following benchmark task and datasets. Where possible, we utilized existing Python libraries such as TorchGeo~\cite{stewart2022torchgeo} and GEO-Bench~\cite{lacoste2024geo}. For a complete list of datasets, please consult \cref{tab:all-ds}. Most datasets are implemented using the TorchGeo library~\cite{stewart2022torchgeo}, when available.

In the following, we outline any modifications we make to standard datasets:

\paragraph{SpaceNet} We utilize the SpaceNet~1 dataset from TorchGeo, which is a building footprint segmentation task over the city of Rio de Janeiro with 8 band MS images and 3 band pansharpened RGB images captured by WorldView 2. We utilize only the 8-band images. Since the original dataset is only available with labels for the training set, we randomly split the dataset into training, validation and test splits with a 80:10:10 ratio.

\begin{table*}
\centering
\resizebox{\textwidth}{!}{%
\begin{tabular}{@{}llll@{}}
\toprule
Index    & Name                  & Name used in this paper & Modifications                                       \\ \midrule
\cir{1}  & Corine [SuperDove]    &                         & Spectral convolution from EnMap to Planet Superdove \\
\cir{2}  & Corine [MODIS]        &                         & Spectral convolution from EnMap to MODIS            \\
\cir{3}  & Hyperview [SuperDove] &                         & Spectral convolution from Intuition to SuperDove    \\
\cir{4}  & Hyperview [MODIS]     &                         & Spectral convolution from Intuition to MODIS        \\
\cir{5}  & TropicalCyclone       & TC                      & TorchGeo, we use 10\% of train                       \\
\cir{6}  & DigitalTyphoon        & DT                      & TorchGeo, we use 10\% of train                       \\
\cir{7} & SpaceNet~1 & SpaceNet~1 & Randomly split the train set into train, val, and test (80:10:10) \\
\cir{8}  & EuroSAT               & m-eurosat               & GEO-Bench                                           \\
\cir{9}  & BrickKiln             & m-brick-kiln            & GEO-Bench                                           \\
\cir{10} & EnMAP-Corine          & Corine                  & Original 202 band dataset                           \\
\cir{11} & RESISC45              & RESISC                  & GEO-Bench                                           \\
\cir{12} & reBEN-S2              &                         &                                                     \\
\cir{13} & reBEN-S1              &                         &                                                     \\
\cir{14} & ForestNet             & m-forestnet             & GEO-Bench                                           \\
\cir{15} & So2Sat                & m-so2sat                & GEO-Bench                                           \\
\cir{16} & PV4Ger (cls.)          & m-pv4ger                & GEO-Bench                                           \\
\cir{17} & PV4Ger (segm.)        & m-pv4ger-seg            & GEO-Bench                                           \\
\cir{18} & Cheasapeake Landcover & m-chesapeake-landcover  & GEO-Bench                                           \\
\cir{19} & Cashew Plantation     & m-cashew-plantation     & GEO-Bench                                           \\
\cir{20} & SA Crop Type          & m-SA-crop-type          & GEO-Bench                                           \\
\cir{21} & NZ Cattle             & m-nz-cattle             & GEO-Bench                                           \\
\cir{22} & NEON Tree             & m-NeonTree              & GEO-Bench                                           \\ 
\cir{23} & fMoW                  &                         & Subset to WV2+WV3 sensors                           \\\bottomrule
\end{tabular}%
}
\caption{Summary of all evaluation datasets and the modifications made to them.}
\label{tab:all-ds}
\end{table*}

\begin{figure*}[!t]  
    \centering
    \includegraphics[width=\textwidth]{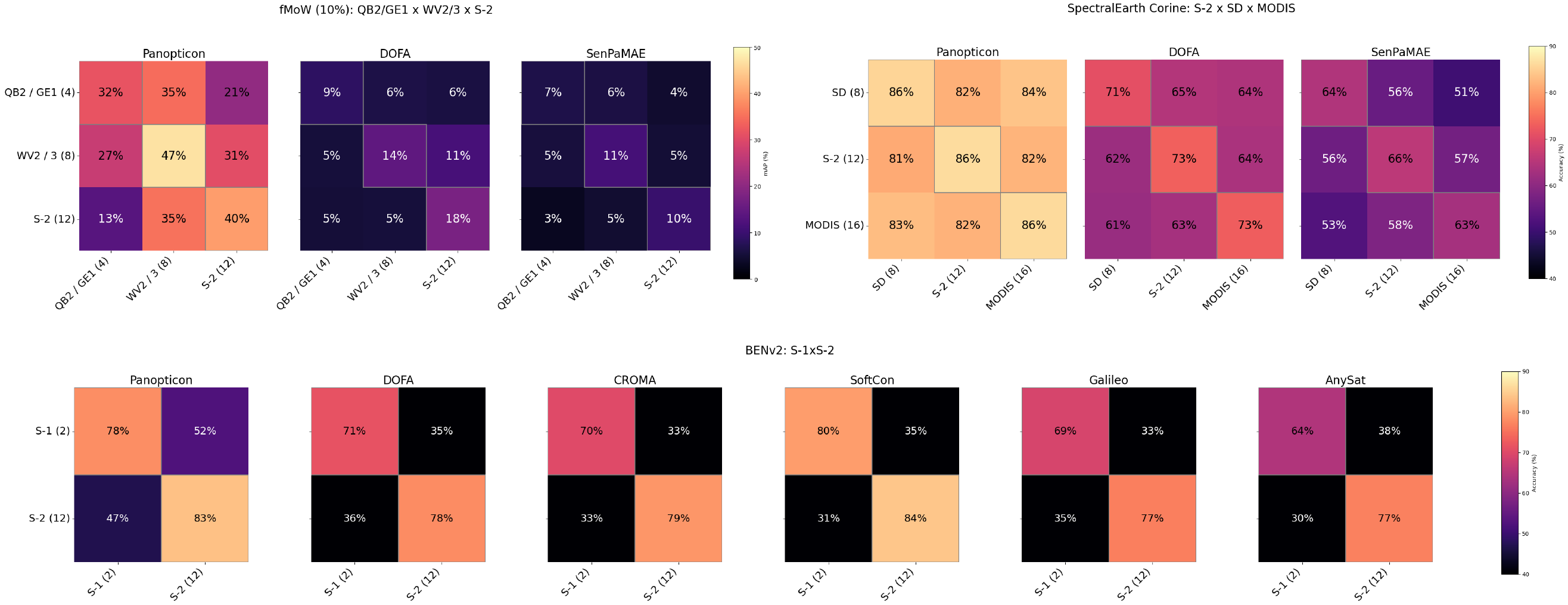}
    \caption{Cross-sensor invariance (absolute values). In addition to train/val/test splits, we also split datasets across sensors to explicitly test sensor invariance. $y$-axes on the heatmap represent training sensors and $x$-axes, test sensors. The diagonals represent same-sensor for train and test, while off-diagonal elements represent cross-sensor results with all values expressed as absolute performance values as percentages.  
    \textbf{Left}: splits are across QuickBird-2 (QB2) / GeoEye-1 (GE1) which are RGB-NIR.
    \textbf{Right}: splits are across synthesized sensors from the HS EnMAP sensor using spectral convolution - MODIS, Sentinel-2 and Planet SuperDove, for any-sensor models. \textbf{Bottom}: reBEN; splits are implemented across Sentinel-1 and Sentinel-2 senors, for all any-sensor and many-sensor models, except SenPaMAE which cannot process SAR imagery. Panopticon consistently outperforms other models in both settings. Note that the value ranges differ for each sub-figure.}
    \label{fig:sup-cross-sens}
\end{figure*}

\section {Additional Results}\label{sec:app_additional_results}
\subsection{Sensor invariance}
\label{sec:cross-sens}
We explicitly test for a model's ability to generate stable representations regardless of the sensor input. To do this, we implement an additional split dimension on datasets that have paired sensors, splitting on the sensor. We use the \cir{10}~EnMAP-Corine~\cite{braham2024spectralearth} dataset where we employ spectral convolution to simulate MODIS, Sentinel-2, and Planet SuperDove sensors (\cref{fig:cross-sens} (left)) and the \cir{12}~reBEN~\cite{clasen2024reben} dataset that has paired imagery from Sentinel-1 and Sentinel-2 (Supp. \cref{fig:cross-sens} (right)). Models are trained on the training split of the dataset using a single sensor, while being validated and tested on a corresponding split of the dataset, using data from a different sensor. We then plot the relative difference to the same-sensor setting as shown on the off-diagonal cells. This allows us to validate how close representations generated from one sensor are to ones generated from a different sensor on the same dataset and prediction objective. Ideally, the off-diagonal cell values are close to 0\%, i.e. similar to the same-sensor setting. This is the case for Panopticon for the Corine dataset in \cref{fig:cross-sens} (left), where other any-sensor models struggle to generalize, especially when training on fewer bands (SD with 8 bands) and testing on sensors with more bands (MODIS with 16 bands). This effect is less pronounced in \cref{fig:cross-sens} (right), where the domain shift is very strong going from optical to SAR and vice-versa. However, even in this setting Panopticon outperforms all any-sensor and many-sensor models by 18\% and 14\%, for the S1$\rightarrow$S2 and S2$\rightarrow$S1 settings on average, respectively.

Absolute values for these experiments along with additional results on the fMoW dataset split according to three included sensors, can be seen in Supp. \cref{fig:sup-cross-sens}.

\begin{figure*}[!t]
    \centering
    \includegraphics[width=1\linewidth]{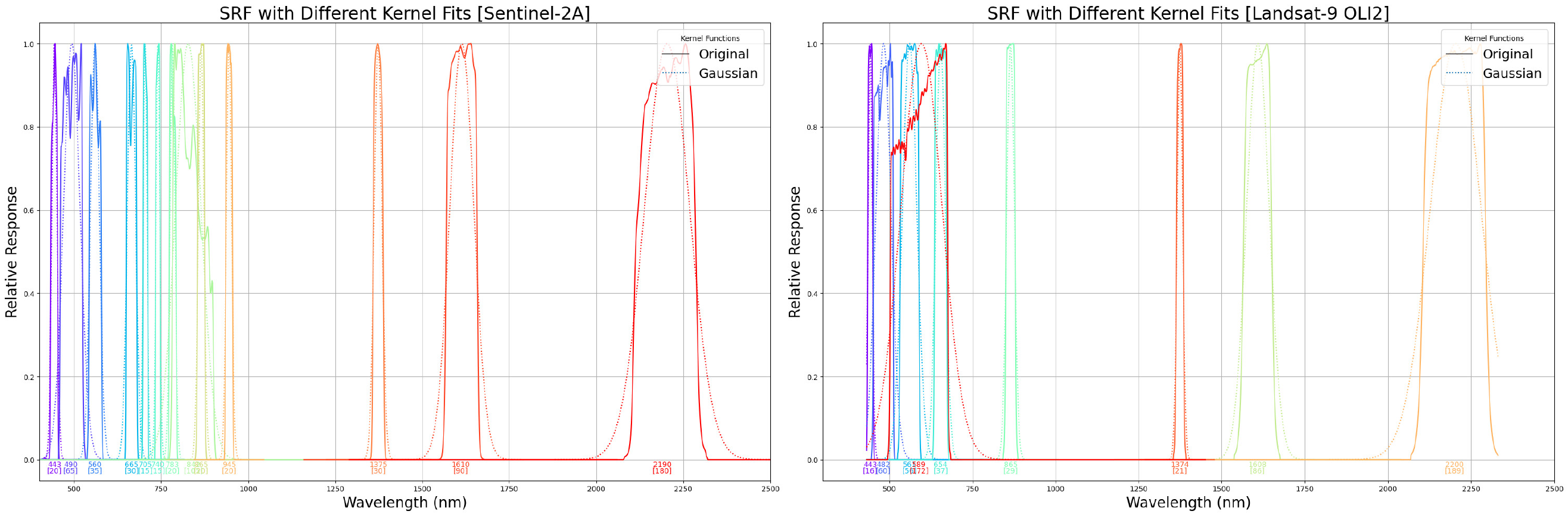}
    \caption{Spectral response function (SRF) fitting for Sentinel-2 (left) and Landsat~9 (right).}
    \label{fig:srf-fitting}
\end{figure*}

\subsection{Geo-Bench}
We present the full results on the Geo-Bench classification datasets in \Cref{tab:gb_cls} and on the segmentation datasets in \Cref{tab:gb_segm_micro_jaccard}, \Cref{tab:gb_segm_macro_jaccard}, and \Cref{tab:gb_segm_miou} reporting different metrics.

\subsection{Influence of individual design decisions}\label{sec:design-dec}
We evaluate the contributions of individual design decisions in \Cref{tab:abl_cumulative}. We identify spectral progressive pre-training and the sample sensor augmentation as the most influential design decisions. Interestingly, removing the DINOv2 weights and training from scratch only results in a performance decrease of 6.1 percentage points in the Avg metric.

\begin{table}[t]
    \centering
    \begin{minipage}[t]{0.48\textwidth}
        \centering
        \begin{tabular}{lrr}
        \toprule
        Individual contributions & \multicolumn{2}{c}{$\text{Avg}\uparrow$} \\
        \midrule 
        Panopticon & $83.5$ &  \\
        $-$ Spectral Progr. Pre-Training & $72.2$ & $-11.3$ \\
        $-$ Sample sensor augm. & $73.2$ & $-10.3$ \\
        $-$ Chn. attn. & $76.7$ & $-6.8$ \\
        $-$ DINOv2 weights & $77.4$ & $-6.1$ \\
        $-$ $D_\text{attn}$ to 2304 & $79.9$ & $-3.6$ \\
        $-$ Spectral cropping & $81.3$ & $-2.2$ \\
        \bottomrule            
        \end{tabular}
    \end{minipage}
    \caption{Impact of individual design choices. We take Panopticon, remove individual design choices, and observe the decrease in performance in the Avg metric. Progr.=progressive, augm.=augmentation, Chn. attn. = channel attention }
    \label{tab:abl_cumulative}
\end{table}

\section{Utilizing complete spectral information}\label{sec:std}

In the field of any-sensor models, DOFA~\cite{xiong2024neural} uses a hypernetwork and the mean of the SRF to learn spectral encodings per channel, while SenPaMAE~\cite{prexl2024senpa} utilizes the full SRF. We experimented with the following mechanisms to incorporate SRF and/or bandwidth information per channel, in addition to the mean wavelength.

\paragraph{Spectral integrated positional encoding}

To achieve sensor agnosticism, we introduce a novel spectral integrated positional encoding (sIPE) that leverages known sensor characteristics. Building on the channel-wise attention mechanism from \citet{nguyen2023climax}, we model each sensor's SRF as an un-normalized Gaussian kernel characterized by its mean, $\mu$ and standard deviation, $\sigma$. This was inspired by noticing that such a Gaussian kernel provides a relatively good fit for SRFs, such as Sentinel-2A and Landsat-9 OLI sensors as shown in \cref{fig:srf-fitting}. We also experimented with Epanechkinov kernel fitting, which provided better $R^2$ results, but since Gaussians are well understood and have closed-form solutions, they tend to be easier to work with. 
Hence, we model the SRF of a channel with the parameters $\{\mu, \sigma\}$ of a Gaussian fit.
We then extend absolute positional embeddings~\cite{vaswani2017attention} to incorporate this spectral information through integration against sinusoidal basis functions
\begin{equation}
    \begin{aligned}
    \text{PE}_{\text{spectral}}(\mu, \sigma, 2i) = \int e^{-\frac{(\mu - \lambda)^2}{2\sigma^2}} \cdot \sin\left(\omega_i \lambda\right)\cdot \text{d}\lambda,\\
    \text{PE}_{\text{spectral}}(\mu, \sigma, 2i+1) = \int e^{-\frac{(\mu - \lambda)^2}{2\sigma^2}} \cdot \cos\left(\omega_i \lambda\right)\cdot \text{d}\lambda,
    \label{eq:sipe}
    \end{aligned}
\end{equation}
where $\omega=\frac{1}{10000^{\frac{2i}{D}}} $ and $i \in (0,D]$. 

We derive the closed-form solution for \cref{eq:sipe} as follows:
\begin{equation}
    \begin{aligned}
    \text{PE}_{\text{spectral}}(\mu, \sigma, 2i) &= \sigma\sqrt{2\pi} \cdot e^{-\frac{\omega_i^2\sigma^2}{2}} \cdot \sin(\omega_i\mu),\\
    \text{PE}_{\text{spectral}}(\mu, \sigma, 2i) &= \sigma\sqrt{2\pi} \cdot e^{-\frac{\omega_i^2\sigma^2}{2}} \cdot \cos(\omega_i\mu).
    \end{aligned}
    \label{eq:sipe_final}
\end{equation}
\cref{eq:sipe_final} allows us to efficiently generate spectral PEs that are added to channel tokens. Our hypothesis was that while the query learns how best to fuse spectral tokens across channels, it can only do so based on the extracted low-level features within those patches. Adding the sIPE to the patch tokens provides a spectral inductive bias to each token relative to its central wavelength and bandwidth, effectively grounding the patch token to its physical capture attributes.

\paragraph{Spectral convolution}
Inspired by \citet{king2024stars}, we employ spectral convolution~\cite{burggraaff2020biases} as a spectral augmentation for HS inputs. Given a source HS image comprised of $i$ channels, $x^s(\lambda)$, and its spectral response function $\text{SRF}_s$, a spectral convolution $R$ is defined as the integral of the product of $x^s$ and $\text{SRF}_s$, normalized by the integral of its SRF. 
To generate arbitrary MS channels with unknown SRFs, we model their SRF, $\text{SRF}_t$ as un-normalized Gaussian kernel with a mean wavelength, $\lambda_t$, and standard deviation, $\sigma_t$. 
Through this process, we are able to generate novel multi-spectral views from any HS source, extensively expanding the spectral augmentation capabilities of this framework. We hypothesize that this combination may prove to add useful spectral inductive biases to the model.

\begin{table}
\centering
\caption{Performance metrics for Panopticon-IPE across various datasets. miJacc = micro Jaccard index}
\label{tab:dataset_metrics}
\footnotesize
\begin{tabular}{lr}
\toprule
\textbf{Dataset} & \textbf{Score} \\
\midrule
\multicolumn{2}{@{}l}{\textit{Classification (Accuracy \%)}} \\
\midrule
m-eurosat & 96.1 \\
m-brick-kiln & 95.8 \\
m-pv4ger-cls & 95.8 \\
RESISC45 & 91.2 \\
m-forestnet & 54.0 \\
\midrule
\multicolumn{2}{@{}l}{\textit{Segmentation (miJacc \%)}} \\
\midrule
m-pv4ger & 95.4 \\
m-nzcattle & 92.8 \\
spacenet1 & 90.3 \\
m-neontree & 79.6 \\
m-chesapeake & 78.0 \\
m-cashew & 59.1 \\
m-sacrop & 52.3 \\
\midrule
\multicolumn{2}{@{}l}{\textit{Multi-Label Classification (mAP \%)}} \\
\midrule
Corine-MODIS & 80.0 \\
Corine-SuperDove & 79.8 \\
\midrule
\multicolumn{2}{@{}l}{\textit{Regression (MSE)}} \\
\midrule
DigitalTyphoon & 0.93 \\
Hyperview-MODIS & 0.35 \\
Hyperview-SuperDove & 0.35 \\
TropicalCyclone & 0.28 \\
\bottomrule
\end{tabular}
\label{tab:sup-sipe-results}
\end{table}

\paragraph{Evaluation}
To evaluate these design choices, we run the evaluation suite on a model trained according to the specifications laid out in \Cref{sec:impl_details}, i.e. identical to the model described in the main paper. We call this model Panopticon-IPE.
The results are shown in \cref{tab:sup-sipe-results}.

Comparing these results to Tab 2 \& 3, 
we see that Panopticon-IPE is relatively close in performance to Panopticon, and in some cases (TropicalCyclone) even excels. However, on average this model performs worse than our default settings, and as a result, we did not include these methods in the main paper. We leave these findings for the benefit of future researchers.

\section{Technical details}\label{sec:app_technical_details}
\subsection{Pre-training}
\paragraph{Implementation of Channel Attention}
We implement the initial shared 2D convolution of the PE with a $1\times p \times p$ 3D convolution~\cite{bao2023channel}, where $p$ is the patch size. Note that images within a batch can originate from different sensors in our pipeline and, thus, the number of channels is not consistent within a batch. To efficiently compute the cross attention for batched inputs, we hence employ padding and masking.

\paragraph{Hyperparameters} 
We follow most of the DINOv2 configuration with the following changes: We multiply the learning rate of the ViT blocks in the backbone by 0.2, reduce the iBOT loss weight to 0.1, and remove the KoLeo regularizer.
We performed early off-the-record ablations on these choices. Apart from that, and for both stages 1 and 2, we train for 70 artificial epochs with 1250 iterations each, with the AdamW optimizer, a base learning rate of 5e-4, standard learning rate scaling $\text{lr}\cdot\text{bsz}/256$, and a linear learning rate warmup for 5 epochs followed by a cosine decay to a minimum 1e-6 learning rate.

\paragraph{Metrics}
For details on the average metrics defined in the ablations, see \cref{tab:abl_metrics}.

\begin{table*}[htbp]
    
    \centering
    \begin{tabular}{ccccc}
        \toprule
         Agg. & Id & Dataset & Task details & Metric \\
         \midrule
          $\text{MS}_\text{acc}$ & 1 & m-eurosat & $k$NN, $k=20$, $T=0.07$ & top-1 micro accuracy \\
          & 2 & \makecell{m-eurosat \\without RGB channels} & $k$NN, $k=20$, $T=0.07$ & top-1 micro accuracy \\

         \hline
          $\text{SAR}_\text{acc}$ & 3 & m-eurosat-SAR \cite{wang2023feature} & $k$NN, $k=20$, $T=0.07$ & top-1 micro accuracy \\
          & 4 & m-eurosat-SAR & linear probing for 10 epochs  & top-1 micro accuracy \\

         \hline
         $\text{Sim}_\text{mAP}$ & 5 & Corine-4  & \makecell{linear probing on \\10\% subset for 10 epochs} & \makecell{top-1 micro \\multilabel average precision} \\
         & 6 & Corine-12 & \makecell{linear probing on \\10\% subset for 10 epochs} & \makecell{top-1 micro \\multilabel average precision} \\

        \hline
        Avg & 1--6 & & & \\
        & 7 & RESISC45 & $k$NN, $k=20$, $T=0.07$ & top-1 micro accuracy \\
        & 8 & m-eurosat only RGB channels & $k$NN, $k=20$, $T=0.07$ & top-1 micro accuracy \\

         \bottomrule
    \end{tabular}
    \caption{Metrics used to inform design decisions and reported in the ablation section of the main text. The aggregation metric is computed as the average of all its metrics. Corine-$n$ denotes the Corine dataset subsampled to $n$ fixed randomly-selected channels. }
    \label{tab:abl_metrics}
\end{table*}

\subsection{Evaluation}
All tasks were executed on either a 40~GB A100 or a 48~GB L40S GPU. The batch size is optimized for each task and model to maximize GPU memory and the base learning rate $lr$ is scaled by the batch size according to the linear scaling rule $\text{lr}\cdot \text{bsz}/256$ \cite{goyal2018accuratelargeminibatchsgd}.
In our evaluations, we use the following tasks types.

\paragraph{$k$NN} $k$-nearest neighbors ($k$NN) with $k=20$ and temperature 0.07 following \citet{reed2023scale}.

\paragraph{Linear probing} We mainly follow the implementation of DINOv2 and sweep multiple extraction methods. In particular, we extract the tokens from the last one or four transformer blocks and aggregate them by concatenating the cls tokens, average pooling, or the default aggregation suggested by the specific model at hand. For each aggregation, we sweep the 13 base learning rates 1e-5, 5e-5, 1e-4, 5e-4, 1e-3, 5e-3, 1e-2, 5e-2, 0.1, 0.5, 1, 5, and 10, resulting in $2\cdot3\cdot13=78$ runs. Note that we use an extension of the DINOv2 implementation that computes all these different configurations simultaneously from a single forward pass of the backbone. 
We train for 50 epochs with a 0.9 momentum Stochastic Gradient Descent optimizer. We also add a trainable batch normalization before the linear head. For the cross-sensor evaluations in \cref{fig:cross-sens} and \cref{fig:sup-cross-sens}, we only extract tokens from the last transformer block to ease compatibility issues across models that use different encoders for SAR and optical modalities.

\paragraph{Linear probing with re-initialized patch embedding} We replace the patch embedding of the model with a randomly initialized 2D convolution layer with the correct number of channels of the dataset at hand. We add a trainable batch norm and linear head to the encoder, unfreeze the new patch embedding and freeze the remaining encoder. 
We fix the feature aggregation to the default one suggested by the model authors and sweep the base learning rates 0.01, 0.001, 0.0005, and 0.0001 with 50 epochs each, stochastic gradient descent optimizer and 5 epochs of learning rate warmup. 

AnySat and Galileo presented unique challenges due to the way they implement modality specific encoders. For AnySat, we pick a specific modality, and retrain the 2D convolution layer of that branch; we picked the NAIP encoder since it does not implement temporal attention, which consumes significant memory and compute. For Galileo, this was not possible since they use a channel grouping where each group produces a set of independent tokens, unlike other models where the channel dimension is collapsed. To replace this, we would have to create a new group with a custom number of channels which would break how Galileo processes tokens in groups. Furthermore, Galileo employs FlexiPatchEmbed, which is trained by randomizing the patch size during pretraining. To properly train this module, we would need to mimic that during evaluation, which was beyond the scope of the evaluation phase of this paper.
Therefore, we omitted evaluating Galileo in tests that required retraining the patch embed module for domain adaptation.

\paragraph{Semantic segmentation} We freeze the backbone and add trainable standard modules from the MMSegmentation library~\cite{mmseg2020}. In particular, we use a Feature2Pyramid as neck, a UPerNet decoder and an auxiliary FCNHead. We sweep the base learning rates 1e-1, 1e-2, 1e-3, 1e-4, 1e-5, 1e-6 and utilize the AdamW optimizer for 50 epochs with no learning rate warmup.

\paragraph{Model adaptations}
During evaluation, we picked the image size that the model was natively trained on to maximize that model's ability to produce representations.

\begin{table*}[!ht]
    \centering
    \begin{tabular}{lrrrrrr}
        \toprule
                  &  m-brick-kiln &  m-eurosat &  m-forestnet &  m-pv4ger &  m-so2sat &  reBEN \\
                 &   (S2)   &  (S2)     &  (L8)      &      (RGB)     &   (S2)        &     (S2)       \\
        \midrule
        DINOv2        &          \textbf{97.5} &       95.5 &         53.5 &      \textbf{97.5} &      60.8 &       80.1 \\
        CROMA         &          94.5 &       91.1 &          - &       - &      53.5 &       79.4 \\
        SoftCon       &          94.9 &       92.2 &          - &       - &      52.1 &       80.6 \\
        AnySat        &          90.3 &       87.6 &         50.9 &      92.8 &       42.5 &        76.8 \\
        Galileo       &          93.1 &       88.6 &          - &       - &      54.2 &       76.5 \\
        SenPaMAE      &          83.9 &       77.5 &         33.5 &      87.1 &      33.7 &        63.8 \\
        DOFA          &          95.8 &       92.9 &         53.2 &      97.4 &      54.2 &       78.8 \\
        Panopticon &          96.7 &       \textbf{96.4} &         \textbf{56.3} &      96.4 &      \textbf{61.7} &       \textbf{83.9} \\
        \bottomrule
        \end{tabular}
    \caption{Linear probing on GEO-Bench classification datasets and reBEN. We report micro accuracy for single-label and mean average precision for multi-label datasets in percentages.}
    \label{tab:gb_cls}
\end{table*}

\begin{table*}[!ht]
    \centering
    \begin{tabular}{lrrrrrr}
        \toprule
         & m-cashew & m-chesapeake & m-neontree & m-nzcattle & m-pv4ger & m-sacrop \\
         & (S2) & (RGBN) & (RGB) & (RGB) & (RGB) & (S2) \\
        \midrule
        DINOv2 & \textbf{65.9} & \textbf{78.5} & \textbf{80.9} & 92.7 & \textbf{96.9} & 51.2 \\
        CROMA & 44.3 & - & - & - & - & 48.4 \\
        SoftCon & 54.5 & - & - & - & - & {51.3} \\
        AnySat & 38.8 & 75.9 & 79.6 & 92.5 & 92.2 & 39.5 \\
        Galileo & 40.4 & - & - & - & - & 39.5 \\
        SenPaMAE & 40.7 & 59.9 & 79.5 & 89.5 & 78.3 & 39.3 \\
        DOFA & 56.4 & 78.2 & {80.4} & \textbf{92.8} & {96.3} & {51.3}\\
        Panopticon & {59.3} & 78.1 & 79.6 & 92.6 & 95.2 & \textbf{52.6} \\
        \bottomrule
    \end{tabular}
    \caption{GEO-Bench segmentation performance. We report tje micro Jaccard index in percentage computed by torchmetrics.classification.JaccardIndex of the torchmetrics package with version x.y.z~\cite{torchmetrics}.
    }
    \label{tab:gb_segm_micro_jaccard}
\end{table*}


\begin{table*}[!ht]
\centering
\begin{tabular}{lrrrrrr}
\toprule
         & m-cashew & m-chesapeake & m-neontree & m-nzcattle & m-pv4ger & m-sacrop \\
         & (S2) & (RGBN) & (RGB) & (RGB) & (RGB) & (S2) \\
\midrule
DINOv2     &  \textbf{53.4} &  \textbf{64.0} &  \textbf{59.1} &  \textbf{76.9} &  \textbf{95.2} &           32.0 \\
CROMA      &           34.3 &        -        &     -           &        -        &         -       &           34.4 \\
SoftCon    &           39.9 &     -           &       -         &      -          &          -      &           33.7 \\
AnySat     &           21.8 &           61.7 &           49.9 &           67.6 &           89.2 &           23.6 \\
Galileo    &           30.4 &       -         &      -          &      -          &       -         &           28.4 \\
SenPaMAE   &           17.8 &           46.9 &           48.3 &           49.6 &           80.4 &           18.4 \\
DOFA       &           41.9 &           59.2 &           55.4 &           74.3 &           94.6 &           32.5 \\
Panopticon &           45.1 &           60.8 &           50.4 &           73.9 &           94.1 &  \textbf{35.7} \\
\bottomrule
\end{tabular}
\caption{GEO-Bench segmentation performance. We report the macro Jaccard index in percentage computed by torchmetrics.classification.JaccardIndex of the torchmetrics package with version 1.6.1~\cite{torchmetrics}}
\label{tab:gb_segm_macro_jaccard}
\end{table*}

\begin{table*}[!ht]
\centering
    \begin{tabular}{lrrrrrr}
\toprule
         & m-cashew & m-chesapeake & m-neontree & m-nzcattle & m-pv4ger & m-sacrop \\
         & (S2) & (RGBN) & (RGB) & (RGB) & (RGB) & (S2) \\
\midrule
DINOv2     &  \textbf{52.7} &  \textbf{26.8} &  \textbf{55.8} &  \textbf{77.4} &  \textbf{92.7} &           13.9 \\
CROMA      &           33.8 &            -    &           -     &      -          &        -        &           14.2 \\
SoftCon    &           39.2 &              -  &      -          &        -        &        -        &           14.2 \\
AnySat     &           20.9 &           25.5 &           47.4 &           67.9 &           86.0 &           10.0 \\
Galileo    &           30.1 &        -        &        -        &        -        &       -         &           11.9 \\
SenPaMAE   &           17.6 &           19.7 &           46.3 &           49.0 &           69.4 &            9.1 \\
DOFA       &           41.2 &           26.6 &           51.5 &           74.6 &           89.7 &           13.9 \\
Panopticon &           44.7 &           26.6 &           47.8 &           73.0 &           88.2 &  \textbf{14.9} \\
\bottomrule
\end{tabular}
\caption{GEO-Bench segmentation performance. We report mIoU averaged over images in percentage computed by torchmetrics.segmentation.MeanIoU of the torchmetrics package with version 1.6.1~\cite{torchmetrics}
}
\label{tab:gb_segm_miou}
\end{table*}


\section{Additional ablations}\label{sec:add_abl}

\paragraph{Spectral augmentation} In line with the DINO ablation on scales in random resized crops, we ablate our spectral size as $(1,s),(s,13)$ for $s\in\{2,4,8,13\}$. In \Cref{tab:cropsplit}, we see that $s=4$ performs best.

\begin{table*}[!htb]
    \centering
    \begin{tabular}{lrrrr}
        \toprule
        $s$ & 2 & 4 & 8 & 13 \\
        \midrule
        $\text{MS}_\text{acc} $ & 81.2 & 85.3 & 83.6 & 81.2 \\
        \bottomrule
    \end{tabular}
    \caption{Ablation of non-overlapping spectral crop size. }
    \label{tab:cropsplit}
\end{table*}
